\documentclass[conference]{IEEEtran}
\IEEEoverridecommandlockouts
\usepackage[T1]{fontenc}
\usepackage{cite}
\usepackage{amsmath,amssymb,amsfonts}
\usepackage{algpseudocode}
\usepackage{algorithm}
\usepackage{algpseudocode}
\usepackage{graphicx}
\usepackage{import}
\usepackage{textcomp}
\usepackage{xcolor}
\def\BibTeX{{\rm B\kern-.05em{\sc i\kern-.025em b}\kern-.08em
    T\kern-.1667em\lower.7ex\hbox{E}\kern-.125emX}}
\begin{document}

\title{Multi-Armed Bandits-Based Optimization of Decision Trees}
\author{\IEEEauthorblockN{Hasibul Karim Shanto}
\IEEEauthorblockA{\textit{Department of Industrial and Production Engineering} \\
\textit{Bangladesh University of Engineering and Technology}\\
Dhaka, Bangladesh \\
hasibsourov36@gmail.com}
\and
\IEEEauthorblockN{Umme Ayman Koana}
\IEEEauthorblockA{\textit{Department of Computer Science} \\
\textit{York University}\\
Ontario, Canada \\
ummekona@yorku.ca}
\and
\IEEEauthorblockN{Shadikur Rahman}
\IEEEauthorblockA{\textit{Department of Computer Science} \\
\textit{York University}\\
Ontario, Canada \\
shadikur@yorku.ca}

}

\maketitle

\begin{abstract}
Decision trees, without appropriate constraints, can easily become overly complex and prone to overfit, capturing noise rather than generalizable patterns. To resolve this problem, pruning operation is a crucial part in optimizing decision trees, as it
not only reduces the complexity of trees but also decreases the probability of generating overfit models. The conventional pruning techniques like Cost-Complexity Pruning (CCP) and Reduced Error Pruning (REP) are mostly based on greedy approaches that focus on immediate gains in performance while pruning nodes of the decision tree. However, this 
might result in a lower generalization in the long run, compromising the robust ability of the tree model when introduced to unseen data samples, particularly when trained with small and complex datasets. To address this challenge, we are proposing a Multi-Armed Bandits (MAB)-based pruning approach, a reinforcement learning (RL)-based technique, that will dynamically prune the tree to generate optimal decision tree with better generalization. Our proposed approach assumes the pruning process as an exploration-exploitation problem, where we are utilizing the MAB algorithms to find optimal branch nodes to prune based on feedback from each pruning actions. Experimental evaluation on several benchmark datasets, demonstrated that our proposed approach results in better predictive performance compared to the traditional ones. This suggests the potential of utilizing MAB 
for a dynamic and probabilistic way of decision tree pruning, in turn optimizing decision tree-based model.
\end{abstract}


\section{Introduction}
According to few studies \cite{castelvecchi2016can} and \cite{rudin2019stop}, one major drawback of most algorithms utilized in machine learning (ML) is their lack of transparency. As machine learning is being applied in all sectors to make things more efficient in recent times, users prefer models that offer explanations about their decision making process. Being a transparent and interpretable algorithm, \cite{blockeel2023decision}, along with its ease of implementation and low computational cost, decision tree based models are becoming more popular day by day in the sector of interpretable machine learning applications \cite{mienye2024survey}. But, a key limitation of decision tree is its tendency to overfit.
This problem of overfitting mainly occurs when the tree becomes too complex and starts to capture noises instead of capturing generalizable patterns in the datasets \cite{schaffer1991does}.

Pruning decision tree is an innovative way of mitigating the effect of overfitting and
improving model's generalization ability \cite{bramer2002pre}. By strategically removing branches that offer minimal contribution to model's predictive performance, pruning not only optimize the decision tree
but also increase its robustness. There are few existing methods for decision tree pruning, such as Reduced Error Pruning and Cost-Complexity Pruning. REP reduces decision tree complexity by removing branches iteratively based on its performance on a validation set, while CCP uses a penalty term to prune the tree while balancing model's complexity and accuracy. Although effective,
they rely heavily 
on heuristic-based criteria or computationally expensive cross-validation procedures. These characteristics limit their scalability, particularly for real world applications.

As pruning a decision tree requires finding best branch node to prune at that moment without hampering the model's predictive performance, it can be designed as a sequential decision making problem. Methods like reinforcement learning shows great potential in this type of decision making tasks. Studies \cite{bertsekas2019reinforcement} and \cite{deliu2024reinforcement} have proven its effectiveness in optimizing sequential decision making tasks. Among various RL framework, Multi-Armed Bandits is really simple and easy to use. It offers sequential decision making in a dynamic way based on iterative feedback, focusing on balancing exploration and exploitation. Unlike current methods for pruning that heavily depend on stiff rules, 
 MAB-based approach can learn an optimal pruning strategy by using feedback from pruning different tree branches to identify the best action.

Furthermore, it could also offer a probabilistic decision making, which can further improve the pruning operation. These characteristics of MAB make it an adaptive decision making framework, which is capable of delivering an efficient and data-driven pruning technique, optimizing decision tree. 

In this paper, we propose a Multi-Armed Bandits-based decision tree pruning framework that dynamically selects branch nodes of decision tree for pruning with an objective of improving model's ability of generalization, interpretability and predictive performance. Our approach formulates the pruning operation as an exploration-exploitation problem where 
each branch node of decision tree is treated as an arm in a bandit framework and pulling an arm represents pruning a branch node.
This approach iteratively updates its pruning action based on feedback from each actions. We evaluate our method on few benchmark datasets and compare its performance against few conventional pruning techniques.



The remainder of this paper is organized as follows:
Section II reviews related work in decision tree pruning, reinforcement learning and multi-armed bandits in optimization and decision trees. Section III outlines 
the proposed MAB-based pruning approach, while Section IV details its evaluation and results. Section V discusses study limitations and Section VI concludes with future research directions.

\section{Related Work}
Pruning operation is important for any network type machine learning algorithms, such as neural network, decision tree etc. It makes the network-based models more optimized while offering better predictive performance. Recently, RL-based methods showed great potential in this sector as they can adapt based on their past actions and strategize for future decisions. In this section, we will review the literature on decision tree pruning, reinforcement learning in optimization and the integration of MAB algorithms for adaptive decision making and development of decision trees.

\subsection{Decision Tree Pruning}
Decision tree pruning is a well established concept, that has been studied extensively over the years and there are various pruning techniques. But all of these techniques can be categorized into two groups. They are:

\begin{itemize}
    \item \textbf{Pre-pruning:} In this strategy, the growth of decision trees are stopped early by enforcing different constraints on decision tree. For example, maximum depth of tree, minimum sample size to create new branch etc. 
    \item \textbf{Post-pruning:} 
     This approach first allows trees to grow fully and then remove branches with low contribution to predictive performance. One big advantage of this technique that it offer better generalized decision tree over pre-pruning approaches. This strategy includes Cost-Complexity Pruning (CCP) \cite{breiman1984cart}, Reduced-Error Pruning (REP) \cite{esposito1997comparative}, Error-Based Pruning \cite{hall2002error} etc. 
\end{itemize}

\subsection{Reinforcement Learning for Model Optimization}
In ML, reinforcement learning has demonstrated advance optimization capability, particularly in the case of hyperparameter tuning, feature selection and neural architecture search. Traditional optimization techniques, such as grid search, Bayesian optimization etc., often struggle with high dimensional search space. As RL-based optimization technique learns from the interactions with the system and uses reward from actions to guide its search, it has proven itself as an alternative of the conventional techniques.

In hyperparameter optimization, \cite{jomaa2019hyp} developed Hyp-RL, a policy based on Q-learning, which can navigate high dimensional hyperparameter spaces. Beyond the optics of hyperparameter optimization, RL has also been used for feature selection process, focusing the improvement of model interpretability as well as computational efficiency. \cite{khurana2018feature} proposed a RL-based feature engineering approach, which finds the most informative features utilizing reward signals. RL has also shown promise in the optimizing network type machine learning models, such as Neural Network, Decision Tree etc. \cite{45826} has proven the effectiveness of reinforcement learning in optimum neural architecture search operation. And for decision tree optimization, studies like \cite{lin2024rl} has demonstrated the performance of RL, leading to better generalization and predictive performance. These achievements highlights the potential of RL for optimizing ML models.

\subsection{Multi-Armed Bandits for Efficient Model Selection}
In optimizing sequential decision making, Multi-Armed Bandits provides an effective optimization mechanism. The most used MAB techniques, Upper Confidence Bound (UCB) algorithm studied in \cite{auer2002finite} and Thompson Sampling studied in \cite{agrawal2012analysis}, are widely adapted for balancing the exploration and exploitation in decision making. In ML,
MAB has been utilized in various tasks. Like RL, 
it has been applied in feature selection purpose, hyperparameter tuning and optimizing neural network structure.
\cite{liu2021multi} has proven MAB's applicability in identifying best set of features for training operation. While \cite{nowakowski2023mab} and \cite{fu2024autorag} display the usefulness of MAB algorithms in machine learning models' hyperparameter tuning. Additionally, \cite{ameen2020pruning} illustrated how MAB techniques can refine the neural network architecture by offering systematic pruning approach. All these studies emphasizes MAB's effectiveness in decision making tasks, aligning with the principles of reinforcement learning. 

\subsection{Multi-Armed Bandits Based Approaches for Decision Trees}
With the increased popularity of decision tree algorithm, several studies have integrated MAB techniques into decision tree learning. The study \cite{tiwari2022mabsplit}, employs MAB algorithm to efficiently find split points during the construction of decision tree and increase the training speed. While \cite{lomax2017cost} leverage MAB techniques to select the best set of attributes during decision tree induction with a goal to generate more cost-effective trees without compromising accuracy. However, these studies focus primarily on the growth of decision trees rather than pruning ineffective branches of those trees.



\subsection{Summary and Research Gap}
Although MAB-based strategies have demonstrated efficiency in model optimization, the study on their application to decision tree pruning still remains an unexplored area. Our work addresses this research gap by introducing a novel adaptive pruning mechanism for decision tree optimization based on Multi-Armed Bandits. It:

\begin{itemize}
    \item Uses MAB-based decision making to identify branch nodes that offer relatively small performance to the model for pruning dynamically.
    \item Balances exploration and exploitation based on reward gained or lost from each action.
    \item Reduces the complexity of decision tree, making it more interpretable and optimized.
\end{itemize}

By leveraging the power of reinforcement learning, MAB-based pruning approach for decision trees could not only offer data-driven optimized decision trees but also improve the predictive performance.

\section{Methodology}
To implement MAB-based pruning of decision tree, first step is to define certain key aspects of MAB. As Multi-Armed Bandits is a simplified version of reinforcement learning, a single state reinforcement learning approach, defining action and reward is the most important part in this algorithm. Here, action means 
a branch node is being replaced with a leaf node in the decision tree, pruning the whole sub-tree from that branch node. To avoid underfitting, only branch nodes beyond level three are considered for pruning.
\begin{figure}[!b]
    \vspace{-10pt}
    \centering
    \includegraphics[height=9cm]{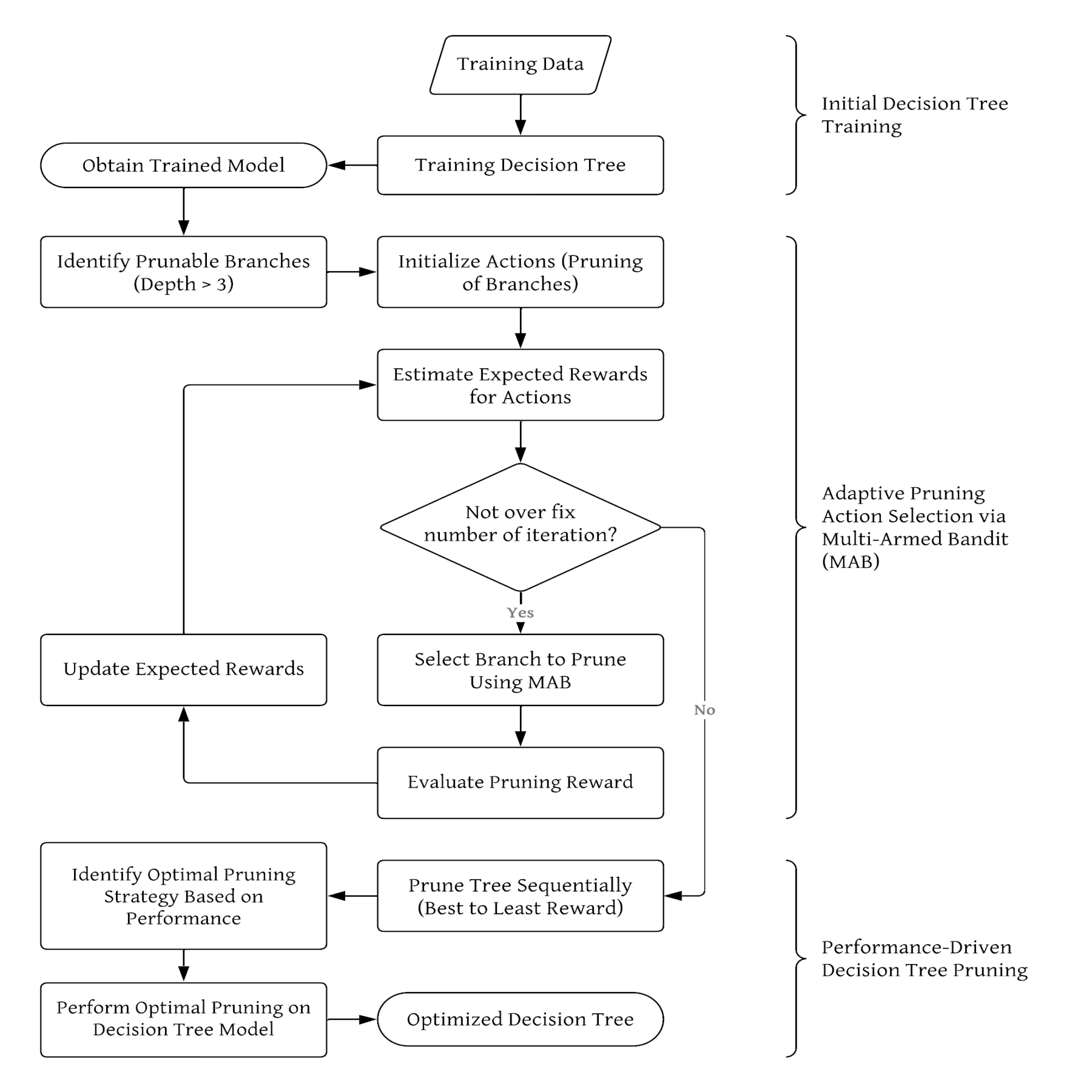} 
    \caption{Overview of the proposed decision tree pruning approach using multi-armed bandit for adaptive pruning and performance-driven optimization.}
    \label{fig:pruning_flowchart}
    \vspace{-10pt}
\end{figure}

\noindent Additionally, the reward is being calculated using a function, utilizing model performance matrices like classification accuracy, logarithmic loss and f1-score. We split the dataset 65:35 into training and testing. Only 0.02\% of the training data is used for consistent evaluation during pruning and for computing rewards, reducing computation compared to using the entire test set for MAB-based optimization. The test data is later being used to assess both pruned and unpruned models.
The actions, selecting branch nodes to be pruned, along with the reward from them are being kept in history and these are later being used by MAB techniques to select future actions. This process is repeated for a fixed number of rounds. Figure 1 illustrates our proposed approach in a flowchart form.

The calculation of reward from pruning, a two step calculation, varies depending on which type of bandits techniques are being used. Although the first step of calculating reward in our proposed approach is similar for all 
techniques:
\begin{IEEEeqnarray}{rCl}
\Delta\text{Score} &=& \alpha(\text{Accuracy}_{\text{new}}-\text{Accuracy}_{\text{old}})
+\gamma(\text{F1}_{\text{new}}-\text{F1}_{\text{old}}) \nonumber\\
&&{}-\beta(\text{Loss}_{\text{new}}-\text{Loss}_{\text{old}})
\label{eq:delta_score}
\end{IEEEeqnarray}

\noindent Here, \(\alpha = 1\), \(\beta = 1\), and \(\gamma = 2.5\) are weight factors controlling the importance of accuracy, scaled logarithmic loss \([0, 1]\) and F1-score, respectively. These weight factor values have been kept consistent throughout our study. The delta score indicates the pruned tree’s performance on a random training subset (0.02\%) relative to the unpruned tree.
However, the second step varies based on the chosen MAB technique.

For example, when we are using Upper Confidence Bound 1 (UCB1), Kullback-Leibler Upper Confidence Bound (KL-UCB) and Win-Stay, Lose-Shift (WSLS) techniques, the reward is being calculated using this function:

\begin{equation}
Reward = \frac{\max(0, \ Threshold+ \Delta Score)}{Constant}
\end{equation}
\noindent where, the value of threshold determines how much loss in performance, negative value of \(\Delta\)Score, is being tolerated after pruning. For example, if pruning results in a less performing tree with a delta value of -0.03, a threshold with a value of 0.05 can still results in a positive reward. Additionally, the value of constant is defined in such way that reward can be bounded between one and zero.

But for Thomson Sampling (TS) and Bayesian Upper Confidence Bounds (Bayes-UCB) techniques, we used a different approach as they follow  Bernoulli rewards, which suggests:

\begin{equation}
Reward = 
\begin{cases} 
1, & if, \ \Delta Score > 0
\\
0, & if, \ \Delta Score \leq 0
\end{cases}
\end{equation}

\subsection{Algorithm}
Given a decision tree model M, number of rounds T, delta function D and reward function R, we have divided the whole 
pruning operation into two steps. Each steps has been displayed as algorithms. Algorithm 1 finds appropriate candidate branch for pruning and Algorithm 2 finds best nodes to prune 
to optimize decision tree.

\begin{algorithm}
\caption{Finding Prunable Decision Tree Branches}
\begin{algorithmic}[1]
\Require Trained decision tree $M$
\Ensure List of prunable branches $B$
\State Initialize list $B \gets \emptyset$
\For{each node $b_i$ in $M$}
    \If{$b_i$ has no children \textbf{and} $\text{depth}(b_i) > 3$}
        \State Add $b_i$ to $B$
    \EndIf
\EndFor
\State \Return List $B$
\end{algorithmic}
\end{algorithm}

\setlength{\textfloatsep}{5pt plus 1.0pt minus 1.0pt}
\begin{algorithm}[!t]
\caption{MAB-Based Decision Tree Pruning}
\label{alg:mab-pruning}
\begin{algorithmic}[1]
\Require Number of rounds $T$, reward function $R$, trained tree $M$, prunable branches $B = \{b_1, \dots, b_n\}$, hyperparameters $\alpha, \beta, \gamma$
\Ensure Optimal pruned tree $M_B$

\State Initialize play count $n_i \gets 0$, average reward $\mu_i \gets 0$ for each $b_i \in B$
\State Evaluate unpruned tree $M$: compute accuracy $A_M$, loss $L_M$, F1-score $f1_M$
\State Set best performance $P_B \gets -100$, best model $M_B \gets M$

\For{$t = 1$ to $T$}
    \State Select a random data subset $S$
    \If{any $b_i$ is untested}
        \State Select $b_i$ randomly
    \Else
        \State Select $b_i$ using MAB policy (e.g., UCB1, Thompson Sampling)
    \EndIf
    \State Temporarily prune $b_i$ to get $M'$
    \State Evaluate $M'$: compute $A_{M'}$, $L_{M'}$, $f1_{M'}$
    \State Compute performance change:
    \[
    D_i = \alpha(A_{M'} - A_M) - \beta(L_{M'} - L_M) + \gamma(f1_{M'} - f1_M)
    \]
    \State Compute reward $R_i = R(D_i)$
    \State Update estimates: 
    \[
    n_i \gets n_i + 1, \quad
    \mu_i \gets \frac{(n_i-1)}{n_i} \mu_i + \frac{1}{n_i}R_i
    \]
    \State Restore $M$
\EndFor

\For{$c = 1$ to $|B|$}
    \State Prune $c$ lowest-ranked branches to obtain $M_c$
    \State Evaluate $M_c$: compute $A_{M_c}$, $L_{M_c}$, $f1_{M_c}$
    \State Compute $P_c = \alpha A_{M_c} - \beta L_{M_c} + \gamma f1_{M_c}$
    \If{$P_c \geq P_B$}
        \State $M_B \gets M_c$, $P_B \gets P_c$
    \EndIf
\EndFor

\State \Return Optimal pruned model $M_B$
\end{algorithmic}
\end{algorithm}

In Algorithm 2, the section of performance evaluation with pruned tree, is dependent on which MAB technique is being used, as different MAB techniques use different approaches of reward calculation. But the calculation of score for each actions taken, later feed into the reward calculation, remains same for all cases. 

\section{Evaluation}
We evaluate our proposed MAB-based pruning approach using five benchmark classification dataset from UCI and Kaggle repositories. The datasets include Breast Cancer Detection \cite{breast_cancer_wisconsin_(diagnostic)_17}, Pima Indian Diabetic Detection \cite{smith1988adap}, Titanic Survival \cite{titanic}, Ionosphere \cite{ionosphere_52} and Credit Card Fraud Detection \cite{dalpozzo2015calibrating}. During the evaluation process, our main focus was to evaluate the usability of the proposed approach against traditional methods used in this purpose.   







\subsection{Experimental Setup}
To ensure consistency, a 
set of hyperparameters, optimized via GridSearchCV, was used to train all decision trees:

\begin{itemize}
\item \textbf{Max Depth = 7:} Balances model complexity and overfitting.
\item \textbf{Min Samples per Leaf = 3:} Ensures reliable predictions at terminal nodes.
\item \textbf{Min Samples per Split = 3:} Prevents overly complex splits.
\end{itemize}

The following 
conditions were uniformly applied for pruning decision trees using
MAB algorithms throughout the study:
\begin{itemize}
\item \textbf{Min Pruning Level:} Nodes above level 3 were excluded to avoid underfitting.
\item \textbf{Min Evaluation Samples:} 0.02\% of training data 
\item \textbf{MAB Iterations:} Set to 1100 to balance exploration and exploitation.
\end{itemize}

As the MAB-based pruning technique relies on iterative evaluation of actions,
the following performance metrics were used to assess the effectiveness of the
pruned models:

\begin{itemize}
\item \textbf{Accuracy:} Measures the generalization capability of the
pruned model on unseen data.
\[
\text{Accuracy} = \frac{TP + TN}{TP + TN + FP + FN}
\]
where TP = True Positives, TN = True Negatives, FP = False Positives, FN = False Negatives.

\item \textbf{Logarithmic Loss (Log Loss):} Evaluates the confidence of probabilistic predictions,
with lower values indicating better-calibrated models.
\[
\text{Log Loss} = -\frac{1}{N} \sum_{i=1}^{N} \sum_{j=1}^{M} y_{i,j} \log(\hat{y}_{i,j})
\]
where, \(N\)= number of samples, \(M\)= number of classes, \(y_{i,j}= 1\) if sample \(i\) belongs to class \(j\), otherwise 0, and \(\hat{y}_{i,j}\)= predicted probability of sample i for class j.

\item \textbf{F1-Score:} Evaluates ML-based models’ predictive performance. It combines both precision and recall scores of any classification models.
\[
\text{F1} = 2 \cdot \frac{\text{Precision} \cdot \text{Recall}}{\text{Precision} + \text{Recall}}
\]
\[
\text{Precision} = \frac{TP}{TP + FP}, \quad
\text{Recall} = \frac{TP}{TP + FN}
\]
\end{itemize}

These three metrics were integrated into a single performance function to guide the MAB pruning process:

\[
\textbf{Performance} = \alpha \cdot \text{Accuracy} - \beta \cdot \text{Log Loss} + \gamma \cdot \text{F1}
\]

where \(\alpha = 1\), \(\beta = 1\), and \(\gamma = 2.5\), as mentioned previously.

\subsection{Empirical Evaluations of MAB-based Pruning Techniques}
The primary goal of this experiment is to evaluate the impact of MAB-based pruning techniques on decision tree performance
against traditional approaches, like CCP, REP etc. We only consider CCP for comparison as it is more sophisticated 
approach compared to REP and both offer similar performance in accuracy as discussed in \cite{wei2005comparison}.

For evaluation, we have applied the proposed technique on five different datasets and evaluated 
their predictive performance by utilizing the scoring function, described in Section IV, on test data. Table~\ref{tab:accuracy_scores} presents the performance scores for each pruning methods on the five datasets. Additionally, Table~\ref{tab:mean_ranks} presents the mean ranking of each method based on their overall performance.\\ 
\begin{table}[t]
    \centering
    \caption{Performance Scores Across Datasets}
    \label{tab:accuracy_scores}
    \resizebox{\linewidth}{!}{ 
    \begin{tabular}{|l|c|c|c|c|c|c|c|c|c|}
        \hline
        Dataset & Unpruned & CCP & UCB1 & TS & SM & WSLS & Bayes-UCB & KL-UCB & Mean Score \\
        \hline
        Diabetics & 1.350 & 1.320 & 1.447 & 1.526 & 1.460 & 1.429 & 1.401 & 1.401 & 1.422 \\
        Credit Card & 2.898 & 3.038 & 3.060 & 3.060 & 3.038 & 3.060 & 3.060 & 3.060 & 3.045 \\
        Breast Cancer & 3.175 & 3.203 & 3.238 & 3.238 & 3.238 & 3.238 & 3.238 & 3.238 & 3.220 \\
        Ionosphere & 2.854 & 2.948 & 3.038 & 3.038 & 3.038 & 3.133 & 3.038 & 3.038 & 3.021 \\
        Titanic & 1.883 & 2.000 & 2.083 & 2.052 & 1.883 & 2.002 & 2.052 & 1.932 & 1.986 \\
        \hline
        Mean Score & 2.432 & 2.502 & 2.573 & 2.583 & 2.571 & 2.572 & 2.554 & 2.554 & 2.543 \\
        \hline
    \end{tabular}
    }
\end{table}
\begin{table}[t]
    \centering
    \caption{Mean Ranks of Different Methods}
    \label{tab:mean_ranks}
    \begin{tabular}{|l|c|}
        \hline
        Method & Mean Rank \\
        \hline
        TS & 2.8 \\
        UCB1 & 2.9 \\
        WSLS & 3.1 \\
        Bayes-UCB & 3.7 \\
        KL-UCB & 4.4 \\
        Softmax & 4.7 \\
        CCP & 6.7 \\
        Unpruned Model & 7.7 \\
        \hline
    \end{tabular}
\end{table}
To provide a formal comparison, 
we conduct an analysis on their statistical significance using paired t-tests, considering the unpruned decision tree and CCP-based pruned tree as baselines. Table~\ref{tab:comparison_before_pruning} and Table~\ref{tab:comparison_ccp} provide a comparative analysis of our proposed method against both unpruned decision tree as well as the CCP-based pruned tree respectively.

\subsubsection{Performance Comparison Against the Unpruned Model}
The  results in Table~\ref{tab:comparison_before_pruning} demonstrate that all six MAB techniques used in the proposed approach outperforms the unpruned decision tree with performance based score improvement in between 4\% and 6.5\%. Among these, Thompson Sampling exhibits the highest score improvement with a value of 6.2\%, which is followed by UCB1 and WSLS, both offering an improvement around 5.8\%.
\begin{table}[t]
    \centering
    \caption{Performance Comparison Against Unpruned Model}
    \label{tab:comparison_before_pruning}
    \begin{tabular}{|l|c|c|c|c|}
        \hline
        Method & Mean Score & Improvement (\%) & T-Statistic & P-Value \\
        \hline
        UCB1 & 2.5732 & 5.81\% & 5.3772 & 0.0058 \\
        TS & 2.5828 & 6.20\% & 6.7770 & 0.0025 \\
        Softmax & 2.5314 & 4.09\% & 3.1337 & 0.0351 \\
        WSLS & 2.5724 & 5.77\% & 3.6315 & 0.0221 \\
        Bayes-UCB & 2.5578 & 5.17\% & 4.4335 & 0.0114 \\
        KL-UCB & 2.5338 & 4.19\% & 3.4658 & 0.0257 \\
        \hline
    \end{tabular}
\end{table}
\noindent To evaluate how these techniques differ in predictive classification statistically, we utilize paired t-tests, where a p-value $< 0.05$ indicates a significant difference between the pair of methods. Thompson Sampling exhibit the lowest p-value $(p=0.0025)$, suggesting a strong statistical significance, followed by UCB1 and Bayes-UCB with statistical significance, $p = 0.0058$ and $p=0.0114$ respectively, depicting their effectiveness compared to unpruned decision tree. Although WSLS shows high level of performance improvement $(5.77\%)$, it's statistical significance $(p=0.0221)$ is slightly lower compared to Bayes-UCB $( p=0.0114)$, which offers lower performance improvement $(5.17\%)$. The rest show significant statistical significance, but their performance improvements vary, while Softmax method shows lowest performance improvement and least statistical significance 4.09\% and 0.0351 respectively.

\subsubsection{Performance Comparison Against Cost-Complexity Pruning}
In this second evaluation, we perform the same tests,
but this time we use the CCP-based pruned tree model as baseline model. The results from this evaluation has been summarized in Table~\ref{tab:comparison_ccp}. It shows different trend from the previous evaluation. Although TS still show highest performance score improvement, its statistical significance is not the most
. Meanwhile, UCB1 provides a well-rounded performance with 2nd highest performance improvement along with the most statistical significance. Here again, the dynamics between WSLS and Bayes-UCB can be seen, as WSLS show higher performance improvement but lower statistical significance then Bayes-UCB. Among all 
, only UCB1 and Bayes-UCB methods have $p<0.05$. Interestingly, Softmax remains the lowest-performing among the six MAB techniques.

\begin{table}[t]
    \centering
    \caption{Performance Comparison Against CCP Method}
    \label{tab:comparison_ccp}
    \begin{tabular}{|l|c|c|c|c|}
        \hline
        Method & Mean Score & Improvement (\%) & T-Statistic & P-Value \\
        \hline
        UCB1 & 2.5732 & 2.85\% & 3.7278 & 0.0203 \\
        TS & 2.5828 & 3.24\% & 2.4343 & 0.0716 \\
        Softmax & 2.5314 & 1.18\% & 0.6770 & 0.5355 \\
        WSLS & 2.5724 & 2.82\% & 2.0865 & 0.1052 \\
        Bayes-UCB & 2.5578 & 2.24\% & 4.2988 & 0.0127 \\
        KL-UCB & 2.5338 & 1.28\% & 1.1359 & 0.3195 \\
        \hline
    \end{tabular}
\end{table}

Overall, based on empirical results, our proposed approach outperforms the baseline models, with UCB1 and Bayes-UCB yielding the most consistent and significant improvements, making them ideal for decision tree pruning.

\section{Limitations}
Although our Multi-Armed Bandits-based pruning approach shows promising results, several limitations should be noted. Firstly, the approach heavily depends on hyperparameter choices such as reward thresholds, accuracy, loss, F1-score weighting, and specific MAB algorithm selection, which might vary across datasets. Secondly, our experiments were limited to small and medium-sized datasets, and its scalability to larger, more complex real-world datasets remains uncertain. Additionally, despite using a smaller data subset for performance evaluation, our reinforcement learning-based approach still has higher computational overhead compared to traditional methods like CCP. Furthermore, we tested only classification tasks; the method's effectiveness on regression or other machine learning tasks has yet to be evaluated. Finally, the stochastic nature of MAB introduces variability, potentially requiring multiple trials to achieve consistent results.

In our future research, we should focus on extensive evaluations on diverse datasets, improved hyperparameter tuning, and extending the approach to other machine learning tasks.

\section{Conclusion}
This study explores the utilization of MAB algorithms for pruning operation of decision tree, focused to not only reduce the complexity of decision tree but also improve predictive accuracy. As decision tree models are quite prone to overfitting, developing an efficient way of generating optimal tree models is essential for better generalization of tree models. Results from our study, conducted on multiple benchmark datasets, indicate that MAB-based pruning techniques perform quite well, particularly UCB1 and Bayes-UCB, because of their all-round performance. As displayed in Table~\ref{tab:accuracy_scores}, all MAB based pruning approaches discussed in our study, outperform the conventional pruning technique, cost-complexity pruning, while maintaining competitive performance across different datasets. Furthermore, the mean rank analysis result, illustrated in Table~\ref{tab:mean_ranks}, indicates the superior performance of MAB-based pruning techniques, suggesting their effectiveness in optimizing decision tree compared to traditional one. This idea is later proven in the t-test analysis, which is exhibited in Table~\ref{tab:comparison_ccp}.

Beyond predictive performance improvement, MAB-based pruning methods offer a probabilistic pruning operation rather than the fixed heuristic approach conventional method based on. This characteristic provide a better data-driven pruning decision which in turns improve the performance of decision tree.

Future research work can explore various extensions of our study. First, integrating contextual bandits into the pruning operation, allowing more adaptive and data-driven decision making for pruning based on contextual information like feature importance, dataset characteristics etc. Second, studying how MAB-based pruning approaches perform in ensemble methods might help to develop more robust and generalized tree-based machine learning models. Lastly, applying MAB-based pruning techniques to real-world application, particularly for medical diagnosis or financial risk assessment where interpretability is necessary, could provide insights about their effectiveness and scalability in complex real-world decision making tasks.  

Developing an efficient pruning technique remains an open challenge, as a decision tree is regarded as a cornerstone of interpretable machine learning algorithms. In this study, we demonstrate a new way of adaptive pruning techniques for decision trees using MAB-based approaches and contribute to the growing research on advancement of interpretable machine learning models.
\bibliographystyle{IEEEtran}
\bibliography{reference}

\begin{thebibliography}{10}
\providecommand{\url}[1]{#1}
\csname url@samestyle\endcsname
\providecommand{\newblock}{\relax}
\providecommand{\bibinfo}[2]{#2}
\providecommand{\BIBentrySTDinterwordspacing}{\spaceskip=0pt\relax}
\providecommand{\BIBentryALTinterwordstretchfactor}{4}
\providecommand{\BIBentryALTinterwordspacing}{\spaceskip=\fontdimen2\font plus
\BIBentryALTinterwordstretchfactor\fontdimen3\font minus \fontdimen4\font\relax}
\providecommand{\BIBforeignlanguage}[2]{{%
\expandafter\ifx\csname l@#1\endcsname\relax
\typeout{** WARNING: IEEEtran.bst: No hyphenation pattern has been}%
\typeout{** loaded for the language `#1'. Using the pattern for}%
\typeout{** the default language instead.}%
\else
\language=\csname l@#1\endcsname
\fi
#2}}
\providecommand{\BIBdecl}{\relax}
\BIBdecl

\bibitem{castelvecchi2016can}
D.~Castelvecchi, ``Can we open the black box of ai?'' \emph{Nature News}, vol. 538, no. 7623, p.~20, 2016.

\bibitem{rudin2019stop}
C.~Rudin, ``Stop explaining black box machine learning models for high stakes decisions and use interpretable models instead,'' \emph{Nature Machine Intelligence}, vol.~1, no.~5, pp. 206--215, 2019.

\bibitem{blockeel2023decision}
H.~Blockeel, L.~Devos, B.~Fr\'enay, G.~Nanfack, and S.~Nijssen, ``Decision trees: from efficient prediction to responsible ai,'' \emph{Frontiers in Artificial Intelligence}, vol.~6, p. 1124553, 2023.

\bibitem{mienye2024survey}
I.~D. Mienye and N.~Jere, ``A survey of decision trees: Concepts, algorithms, and applications,'' \emph{IEEE Access}, 2024.

\bibitem{schaffer1991does}
C.~Schaffer, ``When does overfitting decrease prediction accuracy in induced decision trees and rule sets?'' in \emph{Machine Learning—EWSL-91: European Working Session on Learning Porto, Portugal, March 6--8, 1991 Proceedings 5}.\hskip 1em plus 0.5em minus 0.4em\relax Springer, 1991, pp. 192--205.

\bibitem{bramer2002pre}
M.~Bramer, ``Pre-pruning classification trees to reduce overfitting in noisy domains,'' in \emph{International Conference on Intelligent Data Engineering and Automated Learning}.\hskip 1em plus 0.5em minus 0.4em\relax Springer, 2002, pp. 7--12.

\bibitem{bertsekas2019reinforcement}
D.~Bertsekas, \emph{Reinforcement learning and optimal control}.\hskip 1em plus 0.5em minus 0.4em\relax Athena Scientific, 2019, vol.~1.

\bibitem{deliu2024reinforcement}
N.~Deliu, ``Reinforcement learning for sequential decision making in population research,'' \emph{Quality \& Quantity}, vol.~58, no.~6, pp. 5057--5080, 2024.

\bibitem{breiman1984cart}
\BIBentryALTinterwordspacing
L.~Breiman, J.~Friedman, R.~A. Olshen, and C.~J. Stone, \emph{Classification and Regression Trees}, 1st~ed.\hskip 1em plus 0.5em minus 0.4em\relax Chapman and Hall/CRC, 1984. [Online]. Available: \url{https://doi.org/10.1201/9781315139470}
\BIBentrySTDinterwordspacing

\bibitem{esposito1997comparative}
F.~Esposito, D.~Malerba, G.~Semeraro, and J.~Kay, ``A comparative analysis of methods for pruning decision trees,'' \emph{IEEE Transactions on Pattern Analysis and Machine Intelligence}, vol.~19, no.~5, pp. 476--491, 1997.

\bibitem{hall2002error}
L.~O. Hall, R.~Collins, K.~W. Bowyer, and R.~Banfield, ``Error-based pruning of decision trees grown on very large data sets can work!'' in \emph{14th IEEE International Conference on Tools with Artificial Intelligence, 2002. (ICTAI 2002). Proceedings.}\hskip 1em plus 0.5em minus 0.4em\relax IEEE, 2002, pp. 233--238.

\bibitem{jomaa2019hyp}
H.~S. Jomaa, J.~Grabocka, and L.~Schmidt-Thieme, ``Hyp-rl: Hyperparameter optimization by reinforcement learning,'' \emph{arXiv preprint arXiv:1906.11527}, 2019.

\bibitem{khurana2018feature}
U.~Khurana, H.~Samulowitz, and D.~Turaga, ``Feature engineering for predictive modeling using reinforcement learning,'' in \emph{Proceedings of the AAAI Conference on Artificial Intelligence}, vol.~32, no.~1, 2018.

\bibitem{45826}
\BIBentryALTinterwordspacing
B.~Zoph and Q.~V. Le, ``Neural architecture search with reinforcement learning,'' 2017. [Online]. Available: \url{https://arxiv.org/abs/1611.01578}
\BIBentrySTDinterwordspacing

\bibitem{lin2024rl}
J.~Lin, J.~Zhao, Y.~Deng, Y.~Zhao, W.~Zhou, and H.~Li, ``Rl-llm-dt: An automatic decision tree generation method based on rl evaluation and llm enhancement,'' \emph{arXiv preprint arXiv:2412.11417}, 2024.

\bibitem{auer2002finite}
P.~Auer, ``Finite-time analysis of the multiarmed bandit problem,'' 2002.

\bibitem{agrawal2012analysis}
S.~Agrawal and N.~Goyal, ``Analysis of thompson sampling for the multi-armed bandit problem,'' in \emph{Conference on learning theory}.\hskip 1em plus 0.5em minus 0.4em\relax JMLR Workshop and Conference Proceedings, 2012, pp. 39--1.

\bibitem{liu2021multi}
K.~Liu, H.~Huang, W.~Zhang, A.~Hariri, Y.~Fu, and K.~Hua, ``Multi-armed bandit based feature selection,'' in \emph{Proceedings of the 2021 SIAM International Conference on Data Mining (SDM)}.\hskip 1em plus 0.5em minus 0.4em\relax SIAM, 2021, pp. 316--323.

\bibitem{nowakowski2023mab}
A.~Nowakowski, L.~Str\k{a}k, and W.~Wieczorek, ``Mab-optimized binary pso-based feature selection for enhanced classification performance,'' \emph{Procedia Computer Science}, vol. 225, pp. 4264--4273, 2023.

\bibitem{fu2024autorag}
J.~Fu, X.~Qin, F.~Yang, L.~Wang, J.~Zhang, Q.~Lin, Y.~Chen, D.~Zhang, S.~Rajmohan, and Q.~Zhang, ``Autorag-hp: Automatic online hyper-parameter tuning for retrieval-augmented generation,'' \emph{arXiv preprint arXiv:2406.19251}, 2024.

\bibitem{ameen2020pruning}
S.~Ameen and S.~Vadera, ``Pruning neural networks using multi-armed bandits,'' \emph{The Computer Journal}, vol.~63, no.~7, pp. 1099--1108, 2020.

\bibitem{tiwari2022mabsplit}
M.~Tiwari, R.~Kang, J.~Lee, C.~Piech, I.~Shomorony, S.~Thrun, and M.~J. Zhang, ``Mabsplit: Faster forest training using multi-armed bandits,'' \emph{Advances in Neural Information Processing Systems}, vol.~35, pp. 1223--1237, 2022.

\bibitem{lomax2017cost}
S.~Lomax and S.~Vadera, ``A cost-sensitive decision tree learning algorithm based on a multi-armed bandit framework,'' \emph{The Computer Journal}, vol.~60, no.~7, pp. 941--956, 2017.

\bibitem{breast_cancer_wisconsin_(diagnostic)_17}
W.~Wolberg, O.~Mangasarian, and W.~N. Street, ``{Breast Cancer Wisconsin (Diagnostic)},'' UCI Machine Learning Repository, 1993, {DOI}: https://doi.org/10.24432/C5DW2B.

\bibitem{smith1988adap}
J.~W. Smith, J.~E. Everhart, W.~C. Dickson, W.~C. Knowler, and R.~S. Johannes, ``Using the adap learning algorithm to forecast the onset of diabetes mellitus,'' in \emph{Proceedings of the Symposium on Computer Applications and Medical Care}.\hskip 1em plus 0.5em minus 0.4em\relax IEEE Computer Society Press, 1988, pp. 261--265.

\bibitem{titanic}
W.~Cukierski, ``Titanic - machine learning from disaster,'' \url{https://kaggle.com/competitions/titanic}, 2012, kaggle.

\bibitem{ionosphere_52}
V.~Sigillito, S.~Wing, L.~Hutton, and K.~Baker, ``{Ionosphere},'' UCI Machine Learning Repository, 1989, {DOI}: https://doi.org/10.24432/C5W01B.

\bibitem{dalpozzo2015calibrating}
A.~Dal~Pozzolo, O.~Caelen, R.~A. Johnson, and G.~Bontempi, ``Calibrating probability with undersampling for unbalanced classification,'' in \emph{Proceedings of the IEEE Symposium on Computational Intelligence and Data Mining (CIDM)}.\hskip 1em plus 0.5em minus 0.4em\relax IEEE, 2015.

\bibitem{wei2005comparison}
H.~N. Wei, ``Comparison among methods of decision tree pruning,'' \emph{Journal of Southwest Jiaotong University}, vol.~40, no.~1, pp. 44--48, 2005.

\end{thebibliography}
\end{document}